\documentclass{easychair}
\usepackage{amsfonts} 

\usepackage{booktabs}
\usepackage{doc}
\usepackage{algorithm}
\usepackage[noend]{algpseudocode}
\usepackage{algorithmicx}
\usepackage{amssymb}        

\usepackage{caption}
\usepackage{subcaption}

\usepackage{todonotes}

\usepackage{mathtools}
\DeclareMathOperator*{\argmax}{arg\,max}

\title{Combining Evolutionary Search with Behaviour Cloning for Procedurally Generated Content}

\author{
    Nicholas Muir
\and
    Steven James
}
\institute{
  University of the Witwatersrand
  Johannesburg, South Africa\\
  \email{nicholas.muir1@students.wits.ac.za\\steven.james@wits.ac.za}
 }

\authorrunning{Muir and James}

\titlerunning{PCG with Evolutionary Search and Behaviour Cloning}

\begin{document}

\maketitle

\begin{abstract}
In this work, we consider the problem of procedural content generation for video game levels. 
Prior approaches have relied on evolutionary search (ES) methods capable of generating diverse levels, but this generation procedure is slow, which is problematic in real-time settings. 
Reinforcement learning (RL) has also been proposed to tackle the same problem, and while level generation is fast, training time can be prohibitively expensive. 
We propose a framework to tackle the procedural content generation problem that combines the best of ES and RL.  
In particular, our approach first uses ES to generate a sequence of levels evolved over time, and then uses behaviour cloning to distil these levels into a policy, which can then be queried to produce new levels quickly.
We apply our approach to a maze game and \textit{Super Mario Bros}, with our results indicating that our approach does in fact decrease the time required for level generation, especially when an increasing number of valid levels are required. 
\end{abstract}

\section{Introduction}
\label{sect:introduction}

Procedural content generation (PCG) is used in a wide range of applications, ranging from testing autonomous robot software \cite{arnold2013testing} to the generation of video game levels \cite{ferreira_2014_a}.
With a specific focus on video game design, PCG allows for automatic generation of various aspects of the game, such as level and terrain design \cite{khalifa_2020_pcgrl}. 
As such, it is a powerful tool that can be used by designers to generate more content using fewer resources. 

One of the most common ways of implementing PCG is through the use of evolutionary search, and more specifically genetic algorithms (GAs) \cite{ferreira2014multi}.
GAs are particularly advantageous when it comes to level generation, since they have the ability to create diverse levels.
However, GAs generate their results by intelligently modifying a population of candidate solutions until a suitable one is found.
They are therefore often slow to generate a playable level, which may be problematic when real-time generation is required.

A more recent approach to procedurally generating video game levels is reinforcement learning (RL) \cite{sutton2018reinforcement}, which frames the creation of a level as a sequential decision process and learns a \textit{policy} that produces a sequence of actions to transform an initial randomly generated level into one that is playable and interesting \cite{khalifa_2020_pcgrl}.
Unfortunately, RL requires a reward function specified by a human designer to guide learning, and usually requires significant amounts of training time before a suitable level can be created.
However, after a model has been trained, the generation of levels is fast.  

In this work, we explore the possibility of combining ideas from both ES and RL to inherit the best of both methods. 
In particular, ES is capable of generating multiple diverse levels, while RL is capable of generating a single level quickly.  
We therefore propose an approach that learns a policy using the levels generated by ES, effectively distilling the results of ES into a policy that can be queried to generate new levels quickly. 
Importantly, these policies can be constructed without the extensive training time required by prior work \cite{khalifa_2020_pcgrl}. 

We demonstrate our approach in a Maze game, as well as \textit{Super Mario Bros.}, where results indicate that our approach is able to generate playable levels significantly faster than the competing GA. 

\section{Background}
\label{sec:bg}

In this section, we discuss genetic algorithms and reinforcement learning, two optimisation techniques that have been used in the literature to develop PCG systems. 

\subsection{Genetic Algorithms} 

Genetic algorithms (GAs) seek to optimise a population of candidate solutions. Each individual in the population encodes a particular solution using a genetic representation, or \textit{gene}. 
Individuals are evaluated using a \textit{fitness function}, which is a task-specific function that evaluates the performance of an individual, and is used to decide how they should be modified. 

\begin{figure}[h]
    \centering
    \includegraphics[width=0.8\linewidth]{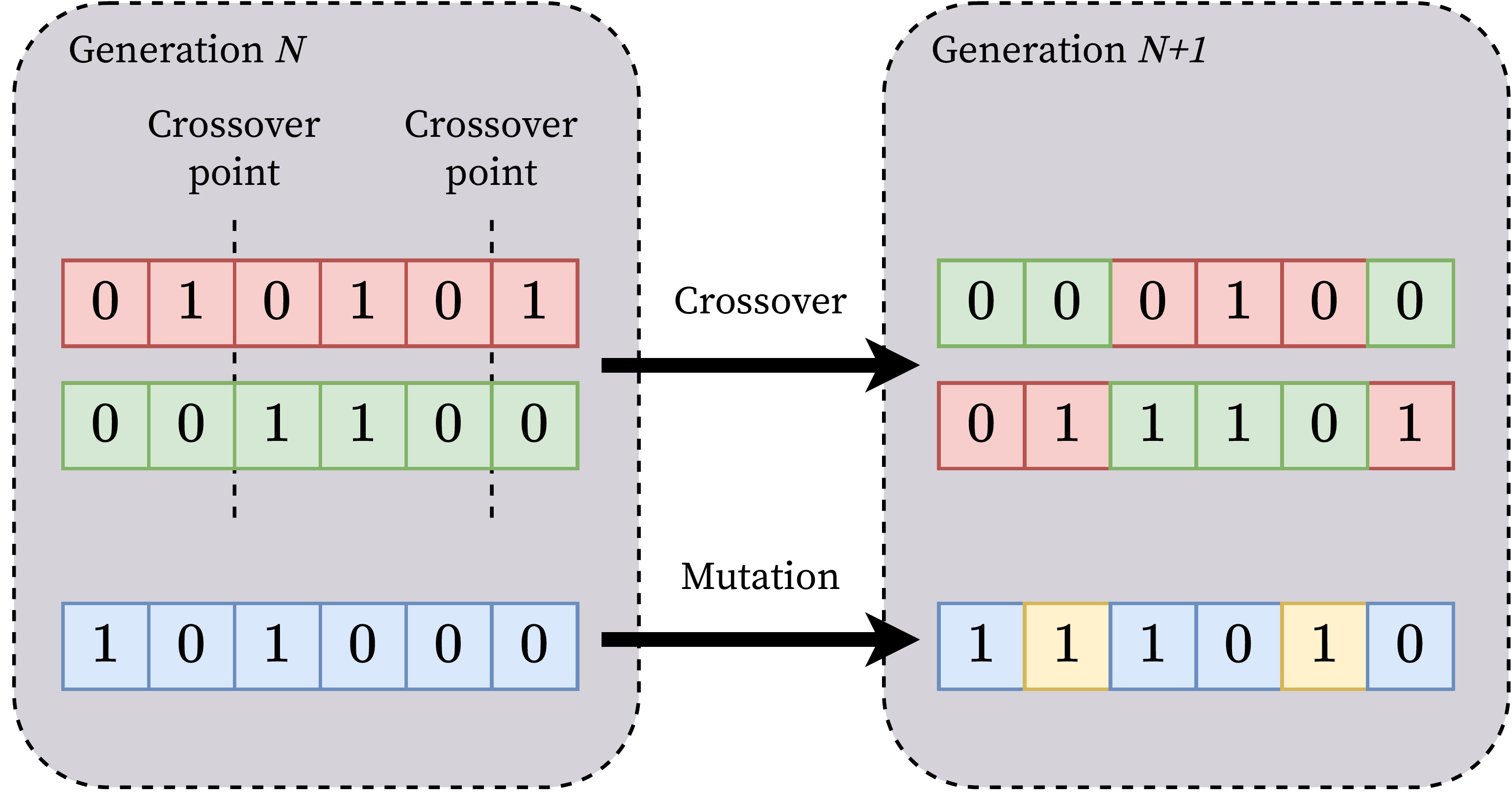}   
    \caption{Illustration of crossover and mutation. Here, solutions are encoded as binary strings. The top row illustrates 2-point crossover, where two genes (red and green) are combined to create two new individuals. The bottom row illustrates mutation, where the genes of the blue individual are perturbed (by flipping a small number of its bits) to produce a new individual.}
    \label{fig:Crossover}
\end{figure}

There are several operators that can be applied to the population in order to increase the fitness of the individuals. 
The most common of these are \textit{crossover}, where two high-performing (according to the fitness function) individuals are combined to form new individuals, and \textit{mutation}, where an individual's gene is perturbed to encourage exploration of the solution space. 
Crossover and mutation are illustrated by Figure~\ref{fig:Crossover}. 

Starting with an initial randomly generated population, GAs modify the population to produce new individuals. Each iteration results in a new \textit{generation} of candidate solutions, which are created using the above operations, and also often inherit the top performing individuals from the previous generation in a concept known as \textit{elitism}. 
The GA continues for a given number of generations, or until individuals surpass some predefined fitness threshold. 

\subsection{Reinforcement Learning}

In reinforcement learning (RL), an agent interacts with an environment in an attempt to solve a given task in a trial-and-error fashion. 
Typically, an environment is modelled as a Markov decision process $\langle S, A, P, R, \gamma \rangle$, where (i) $S$ is the state space; (ii) $A$ is the set of actions available to an agent; (iii) $P(s' | s, a)$ is the transition dynamics, specifying the probability of an agent finding itself in state $s'$ after executing action $a$ from state $s$; (iv) $R(s, a)$ is the reward function that specifies the reward an agent receives for executing action $a$ in state $s$; and (iv) $\gamma \in [0, 1)$ is used to discount future rewards.  

An agent interacts with its environment through a \textit{policy} $\pi$, which maps states to actions. 
The utility of a given policy can be quantified by its \textit{value function}, which captures the expected future rewards following $\pi$:

\[
v_\pi(s) = \mathbb{E}_\pi \left[ \sum\limits_{t=0}^\infty \gamma^t r(s_t, a_t)  | s_t = s \right].
\]
The aim of an agent is to discover an optimal policy $\pi^*$, such that $\pi^* \in \argmax\limits_{\pi} v_\pi(s)$ for all $s$ in $S$. 
This is often achieved through planning approaches such as policy iteration \cite{howard1960dynamic}, or learning approaches such as Q-learning \cite{watkins1989learning}.

\subsubsection{Learning from Demonstration}

An alternate approach to compute a policy relies on an agent having access to the transition data of another agent, consisting of trajectories $\{ s_0, a_0, s_1, a_1, \ldots, a_{n-1}, s_n\}$. 

If these trajectories are generated by an expert, then an agent can use \textit{behaviour cloning} to mimic the expert to learn a direct mapping between states and actions. 
This can be achieved by constructing a dataset $X = \{s_i\}_{i=0}^{n-1}$ with associated labels $Y = \{a_i\}_{i=0}^{n-1}$ and then applying any appropriate supervised learning method to train a model to predict $Y$  given $X$. 
Once trained, this model can then be used directly as a policy, avoiding the need to learn a value function or policy from reward signal alone.

\section{Related Work}

While there have been many approaches to PCG, evolutionary methods are perhaps the most popular \cite{togelius2011search,summerville2018procedural}. 
For example, \cite{ferreira2014multi} use a simple genetic algorithm to generate video game levels, while \cite{liapis2015constrained} use a two-population genetic algorithm. This ensures that a high number of playable levels is generated by maintaining separate populations of feasible (playable) and non-feasible levels. Diverse levels are also encouraged through the use of novelty-based fitness functions \cite{lehman2011abandoning}.  However, game-specific knowledge is also injected to repair unplayable levels, limiting its general applicability.

There has also been work on PCG that leverages the generalisability of neural networks.  For example, \cite{risi2015petalz} represent each collectable game item by a neural network, the weights and structure of which is evolved with a genetic algorithm, while \cite{volz2018evolving,schrum2020cppn2gan} train a generative adversarial network (GAN) on a collection of existing game levels. They
then use evolutionary methods to search for a latent input
vector to this GAN to generate a level. 
However, these approaches require significant amounts of training data and are biassed towards existing levels, which may be undesirable should new and interesting levels be required.

Finally, \cite{khalifa_2020_pcgrl} frame the problem of level generation as a Markov decision process, and use standard RL techniques to learn a policy that generates new levels. Here, actions involve changing a single tile of a 2D map. After to training, generation is fast, and their approach does not rely on prior training data. However, the approach requires a handcrafted reward function, and the training time is extensive (on the order of 100 million timesteps), which is limiting in more complex environments.

\section{Genetic Algorithms with Behaviour Cloning for PCG}
\label{sec:framework}

In Section~\ref{sec:bg}, we described two approaches that have previously been used for PCG.
GAs have been shown to be capable of generating interesting playable levels \cite{ferreira2014multi}, but the search procedure must be executed whenever a new level is required.
By contrast, a policy learned through RL can generate new levels quickly, but the training of such policies is complex and time-consuming. 
Policies can also be computing from demonstration data, but this assumes access to expert trajectories, which is not often the case.

In this section, we describe a novel framework for PCG that addresses the above issues.
The main idea here is to execute a GA to generate levels, and then treat the data generated during the search procedure as the output of an ``expert'' to which behaviour cloning can be applied. 
As a result, our approach produces a policy that can be used to generate new levels quickly, while avoiding the requirement for expensive training or complex, handcrafted reward functions.

In the rest of this paper, we will limit ourselves to 2D tile-based games, where a level is made up of a grid of 2D tiles or cells, each of which takes a specific type (e.g., empty, obstacle, etc). 
Similarly to \cite{khalifa_2020_pcgrl}, an action is represented as the tuple $\langle x, y, t \rangle$, which indicates that the tile at location $(x, y)$ should be modified to type $t$.

\subsection{Phase 1: Evolutionary Search}

\begin{figure}[h]
    \centering
    \includegraphics[width=0.95\linewidth]{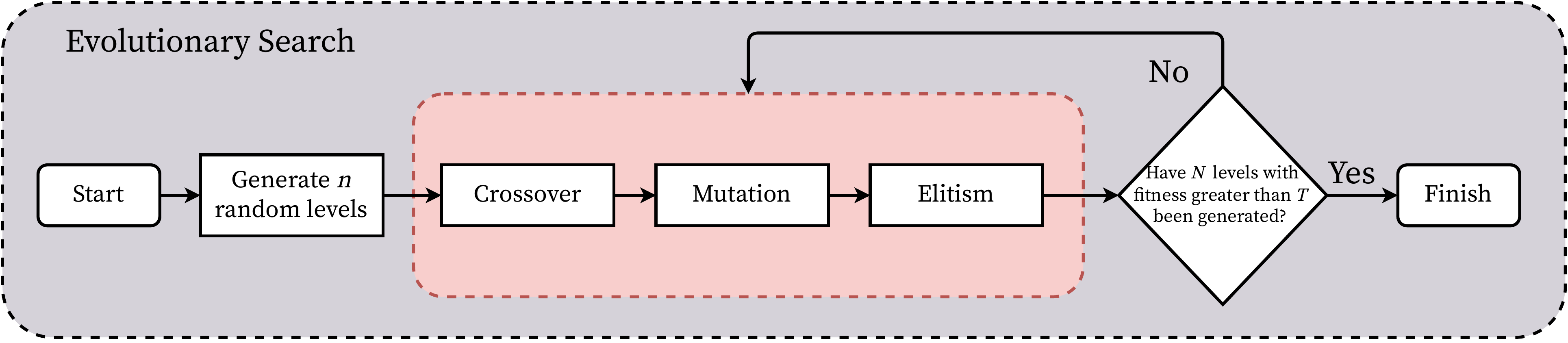}   
    \caption{The process of generating a set of playable levels using a GA. The GA continues until it has produced a sufficient number of levels whose fitness exceeds some threshold.}
    \label{fig:ga}
\end{figure}

We first implement a GA similarly to \cite{safak2016automated} to generate playable levels.
The initial population consists of randomly generated levels, which subsequently undergo both crossover and mutation to create the next generation. 
We also use elitism, retaining a number of high-performing individuals from the previous population. 
After each new generation is formed, we consider the top levels (according to the fitness function) and determine whether they have reached an ``acceptable'' threshold of performance.
If this is the case, the GA terminates, but otherwise continues until a sufficient number of ``acceptable'' levels have been created. 
This process is illustrated by Figure~\ref{fig:ga}.

\subsection{Phase 2: Policy Creation through Behaviour Cloning}

The output of Phase 1 is a set of playable levels, as well as the random levels that were initially created. 
We consider all initial and final levels, and compute the changes in tiles necessary to transform an initial level into the final one.
To generate the data that will be used to fit a policy, we consider each start level $S_i$, end level $E_i$ and set of changes between the two $\Delta_i$. 
Note that because of our action formulation, $\Delta_i$ is equivalently a sequence of actions necessary to transform $S_i$ into $E_i$. 

We then follow the approach in Algorithm~\ref{alg:p2} to generate the data. 
This begins using the state of the initial level, $s_0$ and computing the first action from $\Delta_i$ to apply, $a_0$. The tuple $(s_0, a_0)$ is saved to a buffer, and $a_0$ is applied to $s_0$ to produce a new state $s_1$. This process is repeated until all actions in $\Delta_i$ have been applied. 
All of the above is applied to each start and end level to produce a set of state-action pairs that implicitly represents a policy. 

 \begin{algorithm} 
 \caption{Build policy dataset} \label{alg:p2}
 \begin{algorithmic}
 \Require $ InitialLevels \gets \text{initial randomly generated levels}$
 \Require $ FinalLevels \gets \text{final levels generated by GA}$

\State $Changes \gets \varnothing$

\ForAll{$(S_i, E_i) \in InitialLevels \times FinalLevels$}
\State $s \gets \Call{State}{S_i}$
\State $\Delta \gets \Call{ComputeDiffs}{S_i, E_i}$ \Comment{difference in tiles between start and end levels}
\ForAll{$a \in \Delta$}
\State $Changes \gets Changes \cup \{ (s, a)\}$
\State $s \gets \Call{NextState}{s, a}$
\EndFor
\EndFor
\State $\textbf{Return } Changes$
 \end{algorithmic}
 \end{algorithm}

\subsection{Phase 3: Policy Execution}

To generate a new level, we begin with a randomly created one and must apply our policies, computed in the previous phase, to produce a playable level. 
One issue is that our policy should generalise to unseen states, and while there are many approaches to doing so, here we use a simple approach based on the nearest neighbours algorithm, implemented in \textsc{scikit-learn} \cite{scikit-learn}.

Given an initial randomly generated state, the policy is applied as follows. 
First, the state is passed to the nearest neighbour algorithm to find the most similar state observed in Phase 2.
The action corresponding to this state is then returned and executed.
However, in practice we found that applying a single action to the state does not change it significantly; as a result, the policy may find itself applying the same action in an infinite loop, since the nearest neighbour algorithm continues to return the same state.

We therefore take inspiration from prior work \cite{dabney2020temporally} and execute several actions in sequence before computing the next state. 
This can be seen as executing a temporally extended action for $n$ timesteps. 
Since the buffer created in the previous phase is sequential, we simply execute action $a_i$, and then subsequently $a_{i+1}, \ldots a_{i+n}$.
In practice, instead of using a fixed value of $n$, we instead use hyperparameter $p$, which represents the proportion of the total changes that should be made, $|\Delta_i|$, and compute $n = |\Delta_i| / p$.

\begin{figure*}[h]
    \centering
    \includegraphics[width=0.95\linewidth]{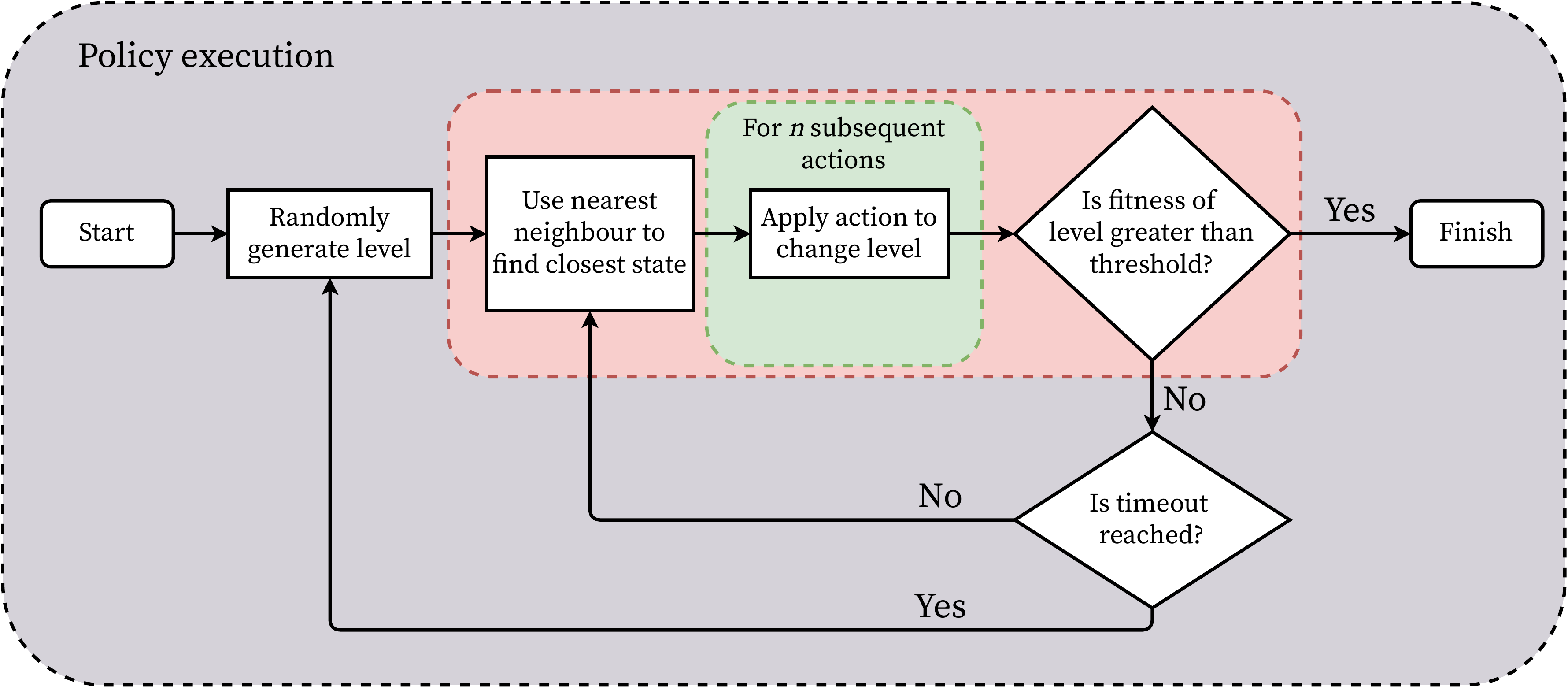}  
    \caption{Applying our policy to transform an initial random level into a playable one. }
    \label{fig:RL}
\end{figure*}

The above process is repeated until one of two conditions are met: either a maximum number of steps is reached, or an acceptable level (according to the fitness function) is generated. In the former case, the algorithm restarts with a new random level. 
This entire procedure is illustrated by Figure~\ref{fig:RL}.

\section{Experiments}

We apply our framework to two tile-based environments: a Maze domain, where an agent is required to navigate from the top left to the botton right of a maze consisting of free space and walls, and the video game \textit{Super Mario Bros}.
We use the \textsc{amidos2006 Mario AI} framework\footnote{\url{https://github.com/amidos2006/Mario-AI-Framework}} to generate \textit{Super Mario Bros} levels of size $101 \times 16$ tiles and to evaluate the fitness of these levels.

\subsection{Maze Domain}

We use the Maze domain as a test-bed with which to validate our framework. A Maze level is considered ``playable'' if there exists a valid path from the top left to the bottom right corner. 
Maze levels are generated by randomly placing walls within the bounds of the maze, and a given level is represented by the $xy$-coordinates of each wall.  

\paragraph{Fitness function} The fitness function for the generated mazes takes into account whether or not the level is playable, the spread of blocks within the maze, and the length of the shortest path from the start to the goal. The first component ensure the maze is solvable, while the second favours mazes with an equal spread of vertical and horizontal walls. Finally, longer optimal solutions are given higher fitness to promote difficult, interesting mazes. 
The exact fitness function is

\begin{equation*}
    f(x) = 0.7 \times Finishable + 0.2 \times RatioX + 0.2 \times RatioY + 0.0001 \times PathLength
\end{equation*}
where  (i) $Finishable$ is $1$ if a solution exists, and $0$ otherwise; (ii) $RatioX$ is the number of walls in the left half of the level divided by the total number of walls;  (iii) $RatioY$ is the number of walls in the top half of the level divided by the total number of walls; and (iv) $PathLength$ is the length of optimal path.
We apply the framework described in Section~\ref{sec:framework} to the Maze domain with the hyperparameters and experimental settings described in Table~\ref{tab:1}.

\begin{table}[h]
\centering
\begin{tabular}{@{}lp{90mm}l@{}}
\toprule
Hyperparameter          & Description                                                                & Value              \\ \midrule
Initial population size & Number of levels to be generated and evaluated for initial batch of levels & $50$ \\
Size of child list      & Size of list of levels with the best fitnesses to determine whether to terminate      & $20$                   \\
Crossover points        & Number of segments each chromosome should be split for crossover           & $50$                    \\
Mutation rate           & Percentage of chromosome to mutate                                         & $0.05$                   \\
Maximum iterations      & Maximum number of iterations for genetic algorithm                         & $1000$                    \\
$p$          & Parameter that controls the length of the temporally extended action & $0.06$                   \\
$D$   & Size of the mazes  & $10$ -- $50$ \\
Number of walls         & Number of walls for in each level & $0.15D^2$\\
\bottomrule
\end{tabular}
\caption{Maze domain hyperparameters and experimental settings.}
\label{tab:1}
\end{table}

\begin{figure*}[h]
\centering
    \centering\includegraphics[width=3.5cm,height=3.5cm]{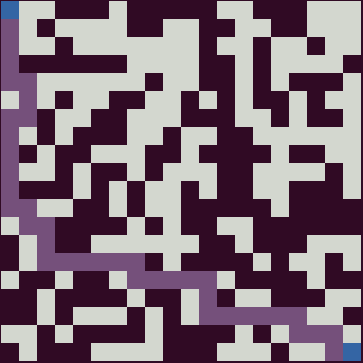}  
    \hspace{20px}
    \centering\includegraphics[width=3.5cm,height=3.5cm]{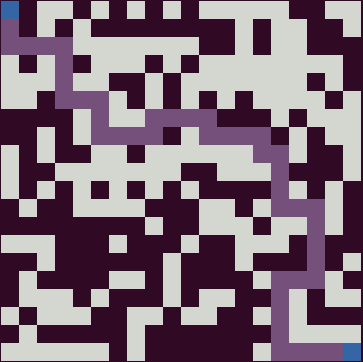} 
    \hspace{20px}
    \centering\includegraphics[width=3.5cm,height=3.5cm]{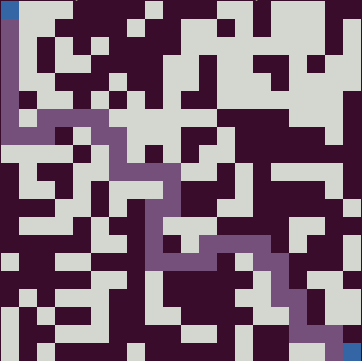}
    \caption{Maze levels of size $20 \times 20$ generated by a genetic algorithm. Purple indicates the optimal path from the start to the goal location.}
    \label{fig:MazeES}
\end{figure*}

\begin{figure*}[h]  
\centering
    \centering\includegraphics[width=3.5cm,height=3.5cm]{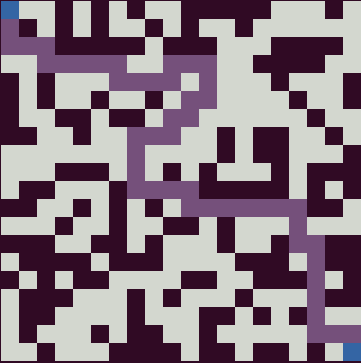}  
    \hspace{20px}
    \centering\includegraphics[width=3.5cm,height=3.5cm]{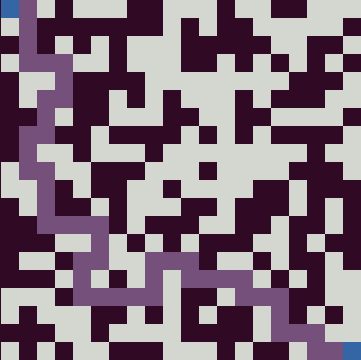}
    \hspace{20px}
    \centering\includegraphics[width=3.5cm,height=3.5cm]{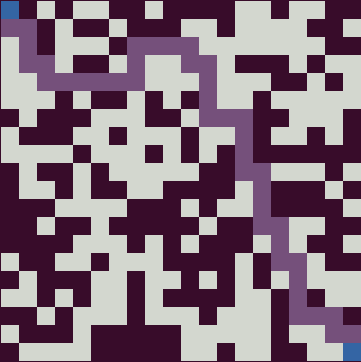}
    \caption{Maze levels of size $20 \times 20$ generated by our approach. Purple indicates the optimal path from the start to the goal location.}
    \label{fig:MazeRL}
\end{figure*}

We evaluate our approach against a genetic algorithm to determine the time taken to generate the same number of playable levels. 
We first note that even though the policy is created from levels generated by a GA, the final levels differ between the two approaches. 
Figures \ref{fig:MazeES} and \ref{fig:MazeRL} illustrate a subset of these levels. 

\begin{figure}[h!]
     \centering
     \begin{subfigure}[b]{0.48\textwidth}
         \centering
         \includegraphics[width=\textwidth]{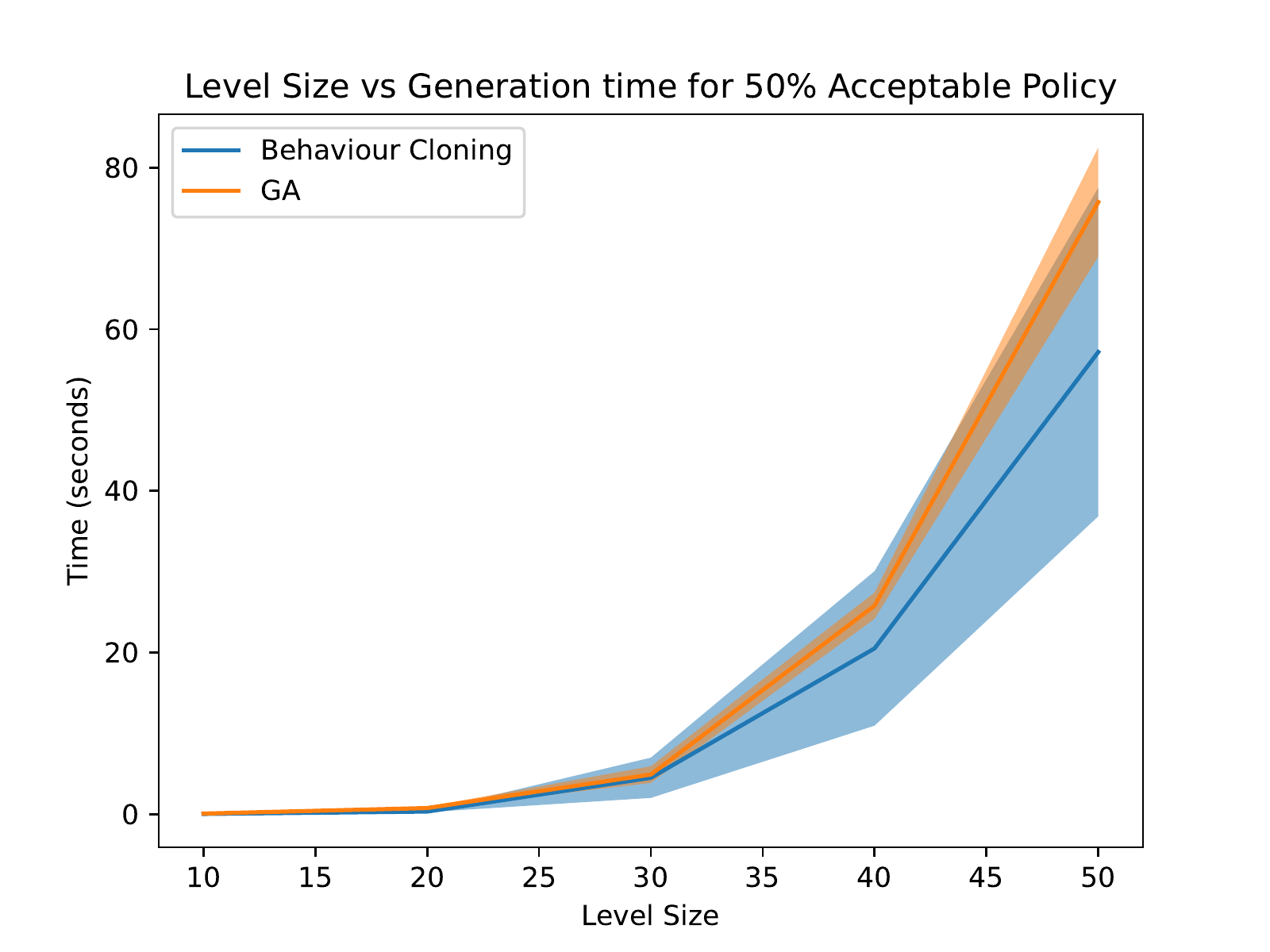}
         \caption{Maze generation with 50\% acceptable levels produced by the GA.}
         \label{fig:ResultMaze50}
     \end{subfigure}
     \hfill
     \begin{subfigure}[b]{0.48\textwidth}
         \centering
         \includegraphics[width=\textwidth]{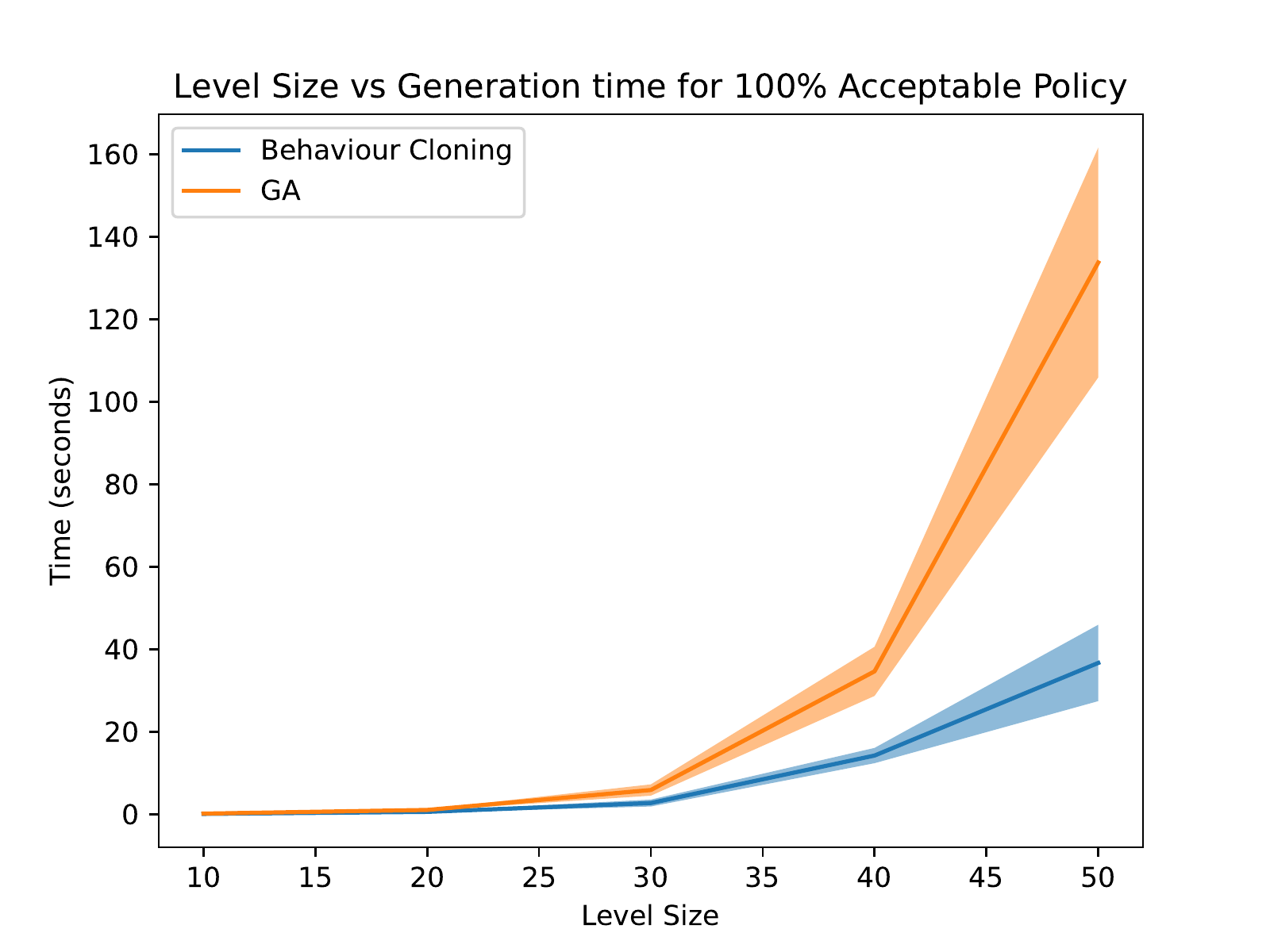}
         \caption{Maze generation with 100\% acceptable levels produced by the GA. }
         \label{fig:ResultMaze100}
     \end{subfigure}
        \caption{A wall-clock comparison of our approach with a genetic algorithm on the Maze domain of various sizes. Lower is better. Mean and standard deviation over 10 random seeds are shown.}
        \label{fig:ResultMaze}
\end{figure}

Our initial tests indicated that applying behaviour cloning to the output of a GA that terminates when at least one individual passes the threshold resulted in poor performance. 
However, when at least half of the GA's individuals were deemed acceptable (i.e. 50\% of the final generation had a fitness greater than the threshold), applying our approach resulted in good performance.
Moreover, when requiring that the final generation of the GA consist of only acceptable levels, our approach vastly outperforms it, especially as the size of the Maze level increases.
Figure \ref{fig:ResultMaze} illustrates that increasing the quality of the GA's output has a positive effect on our framework. 
The results also indicate that our method is able to better scale to larger domains when compared to GAs. 

\subsection{\textit{Super Mario Bros} Domain}

\begin{figure}[h]
    \centering
    \includegraphics[width=0.99\linewidth]{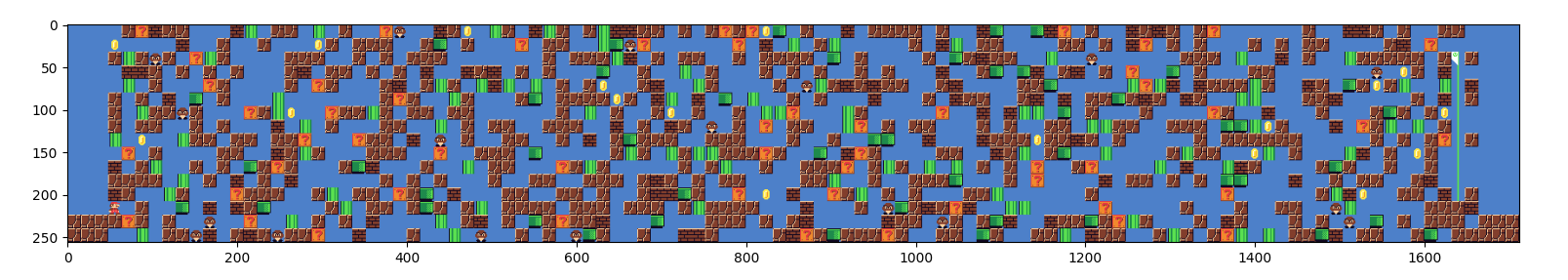}
    \caption{Randomly generated \textit{Super Mario Bros} level.}
    \label{fig:Mariornd}
\end{figure}

We now apply our framework to a significantly more challenging video game environment. 
In \textit{Super Mario Bros.}, the agent traverse the level from left to right while avoiding enemies and obstacles. 
Since there are multiple tile types, each level is represented by the $xy$-position of each tile, as well as its specific type.
To generate a random level, we randomly place tiles within the level, but assign higher probability to ``air'' tiles (which can be passed through) to assist the GA in producing playable levels.
A randomly generated level is illustrated by Figure~\ref{fig:Mariornd}.

\paragraph{Fitness function} For \textit{Super Mario Bros}, we implement a fitness function that includes whether the agent timed out, won or lost the current level (based on behaviour generated by \textsc{amidos2006 Mario AI}). The fitness function also includes the percentage of the level the agent was able to complete before the end state, and the agent's state (whether the agent acquired upgrades within the level). 
Specifically, the fitness function is 

\begin{equation*}
    f(x) = WinState + 2 \times CompletionPercentage + 0.5 \times MarioState ,
\end{equation*}
where  (i) $WinState$ takes the value $0.1$ if a timeout occurs, $0.4$ if the agent loses and $1$ if it succeeds; (ii) $CompletionPercentage$ is the percentage of the screen the agent is able to traverse to the right; and (iii) $MarioState$ is $1$ if any power-ups were collected, and $0$ otherwise.

We apply or approach to \textit{Super Mario Bros} with the hyperparameters and experimental settings described in Table~\ref{tab:2}.

\begin{table}[h]
\begin{tabular}{@{}lp{90mm}l@{}}
\toprule
Hyperparameter          & Description                                                                & Value              \\ \midrule
Initial population size & Number of levels to be generated and evaluated for initial batch of levels & $100$ \\
Size of child list      & Size of list of levels with the best fitnesses to determine whether to terminate      & $20$                    \\
Crossover points        & Number of segments each chromosome should be split for crossover           & $101$                    \\
Mutation rate           & Percentage of chromosome to mutate                                         & $0.05$                   \\
Maximum iterations      & Maximum number of iterations for genetic algorithm                         & $1000$                    \\
$p$          & Parameter that controls the length of the temporally extended action & $0.05$                   \\ \bottomrule
\end{tabular}
\caption{\textit{Super Mario Bros} hyperparameters and experimental settings.}
\label{tab:2}
\end{table}

\begin{figure*}[h]
    \centering
    \includegraphics[width=0.99\linewidth]{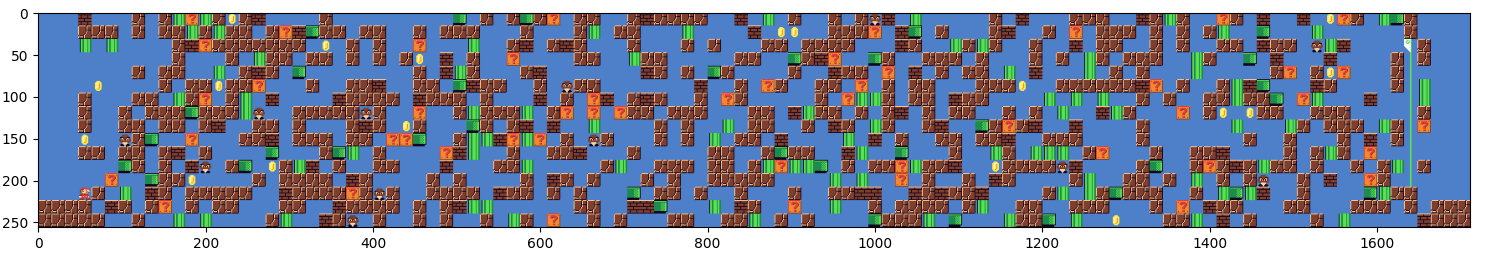}   
    \caption{\textit{Super Mario Bros} level generated by a genetic algorithm.}
    \label{fig:MarioESworks}
\end{figure*}

\begin{figure*}[h]
    \centering
    \includegraphics[width=0.99\linewidth]{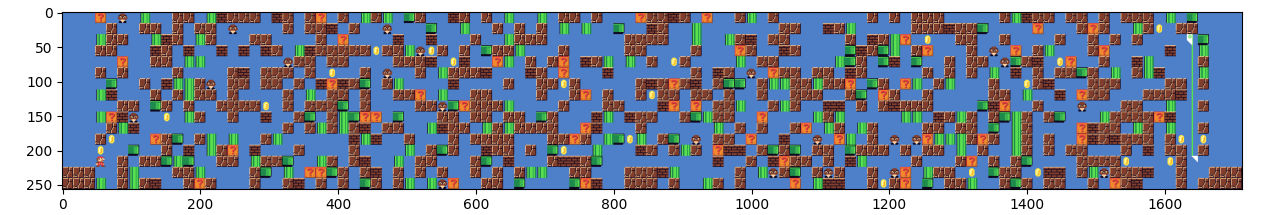}   
    \caption{\textit{Super Mario Bros} level generated by our approach.}
    \label{fig:MarioRLworks}
\end{figure*}

As in the Maze domain, we again note that there is a large difference between the output of the GA and our approach, despite the latter being trained on levels produced by the former. 
This is illustrated by Figures~\ref{fig:MarioESworks} and \ref{fig:MarioRLworks}.

\begin{figure}[h]
    \centering
    \includegraphics[width=0.65\linewidth]{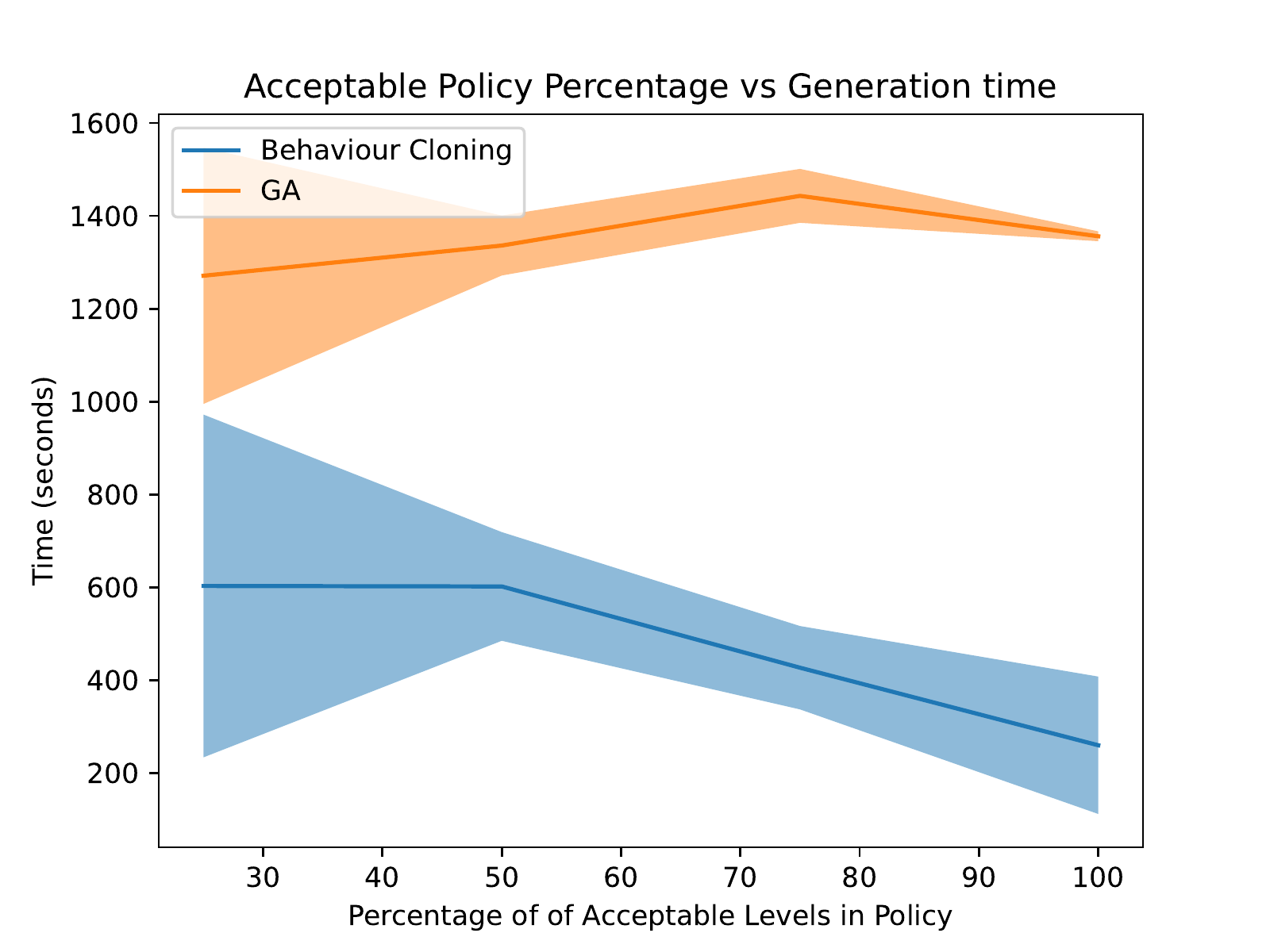} 
    \caption{A wall-clock comparison of our approach with a genetic algorithm on the \textit{Super Mario Bros} domain as a function of the number of levels required by the GA to be acceptable.  Lower is better. Mean and standard deviation over 10 random seeds are shown.}
    \label{fig:MarioRes}
\end{figure}

Finally, we quantitatively compare our approach to a GA to measure how long it takes to generate a given number of playable levels. 
The results in Figure~\ref{fig:MarioRes} clearly indicate that once a policy has been trained on the output of a GA, the creation of subsequent levels is significantly faster than repeatedly executing the GA to produce more levels.

\section{Conclusion}

We have proposed an approach to procedural content generation for video games that relies on a combination of evolutionary search and behaviour cloning. 
Our approach allows an agent to derive a policy capable of generating new levels quickly, without the need for expensive training or complex, handcrafted reward functions. 
Our results on two domains, including a complex video game, indicate that our approach outperforms genetic algorithms.

Our framework is agnostic to the exact details of both the genetic algorithm and the method of behaviour cloning. 
We adopted a simple, na\"{i}ve approach here, but promising future work would be to incorporate more sophisticated algorithms and techniques, such as deep neural networks and more advanced search strategies. 
More generally, the combination of evolutionary search with reinforcement may be a productive avenue for deploying PCG in the real world.  

\newpage

\label{sect:bib}
\bibliographystyle{plain}
\bibliography{easychair}

\end{document}